\documentclass[runningheads]{llncs}
\usepackage{graphicx}
\usepackage{amsmath,amssymb} 
\usepackage{url}

\usepackage{breakurl}
\usepackage[breaklinks]{hyperref}
\usepackage{array}
\usepackage{float}
\usepackage{booktabs}
\usepackage{comment}
\usepackage{subcaption}
\usepackage{color}
\usepackage{flushend}
\usepackage{epstopdf}
\usepackage{multirow}
\usepackage{bm}
\usepackage{paralist}
\usepackage{color}
\usepackage[width=122mm,left=12mm,paperwidth=146mm,height=193mm,top=12mm,paperheight=217mm]{geometry}
\usepackage{hyperref}

\newcommand{\Outer}{Element-wise}
\newcommand{\EleAttGn}{{EleAttG}}
\newcommand{\EleAttG}{{EleAttG~}}
\newcommand{\EARNN}{EleAtt-RNN}

\begin{document}

\renewcommand{\labelitemi}{$\bullet$}
\renewcommand{\labelitemii}{$\cdot$}
\renewcommand{\labelitemiii}{$\diamond$}
\renewcommand{\labelitemiv}{$\ast$}
\newcommand*\samethanks[1][\value{footnote}]{\footnotemark[#1]}

\pagestyle{headings}
\mainmatter

\title{Adding Attentiveness to the Neurons in Recurrent Neural Networks} 

\titlerunning{Adding Attentiveness to the Neurons in Recurrent Neural Networks}

\authorrunning{Zhang et al.}


\author{Pengfei Zhang{\small $^{1}$}, Jianru Xue{\small $^{1}$}\thanks{Corresponding author.}, Cuiling Lan{\small $^{2}$}\samethanks[1], Wenjun Zeng{\small $^{2}$}, Zhanning Gao{\small $^{1}$}, Nanning Zheng{\small $^{1}$}}


\institute{{\small $^{1}$}Institute of Artificial Intelligence and Robotics, Xi'an Jiaotong University, {\small $^{2}$}Microsoft Reserach Asia\\
	\email{ zpengfei@stu.xjtu.edu.cn, \{culan,wezeng\}@microsoft.com, 	\{jrxue,nnzheng\}@mail.xjtu.edu.cn, zhanninggao@gmail.com} 
}

\maketitle

\begin{abstract}
	Recurrent neural networks (RNNs) are capable of modeling the temporal dynamics of complex sequential information. However, the structures of existing RNN neurons mainly focus on controlling the contributions of current and historical information but do not explore the different importance levels of different elements in an input vector of a time slot. We propose adding a simple yet effective \Outer-Attention Gate (\EleAttGn) to an RNN block (e.g., all RNN neurons in a network layer) that empowers the RNN neurons to have the attentiveness capability. For an RNN block, an \EleAttG is added to adaptively modulate the input by assigning different levels of importance, {\it i.e.}, attention, to each element/dimension of the input. We refer to an RNN block equipped with an \EleAttG as an EleAtt-RNN block. Specifically, the modulation of the input is content adaptive and is performed at fine granularity, being element-wise rather than input-wise. The proposed \EleAttGn, as an additional fundamental unit, is general and can be applied to any RNN structures, {\it e.g.}, standard RNN,
	Long Short-Term Memory (LSTM), or Gated Recurrent Unit (GRU). We demonstrate the effectiveness of the proposed EleAtt-RNN by applying it to the action recognition tasks on both 3D human skeleton data and RGB videos. Experiments show that adding attentiveness through EleAttGs to RNN blocks significantly boosts the power of RNNs.

\end{abstract}

\section{Introduction}

In recent years, recurrent neural networks \cite{lipton2015critical}, such as standard RNN (sRNN), its variant Long Short-Term Memory (LSTM) \cite{hochreiter1997long}, and Gated Recurrent Unit (GRU) \cite{cho2014learning}, have been adopted to address many challenging problems with sequential time-series data, such as action recognition \cite{du2015hierarchical}, machine translation \cite{cho14}, and image caption \cite{vinyals2015show}. They are powerful in exploring temporal dynamics and learning appropriate feature representations.

\begin{figure}[!t]
	\centering
	\begin{subfigure}[t]{0.43\linewidth}
		\centering\includegraphics[width=\textwidth]{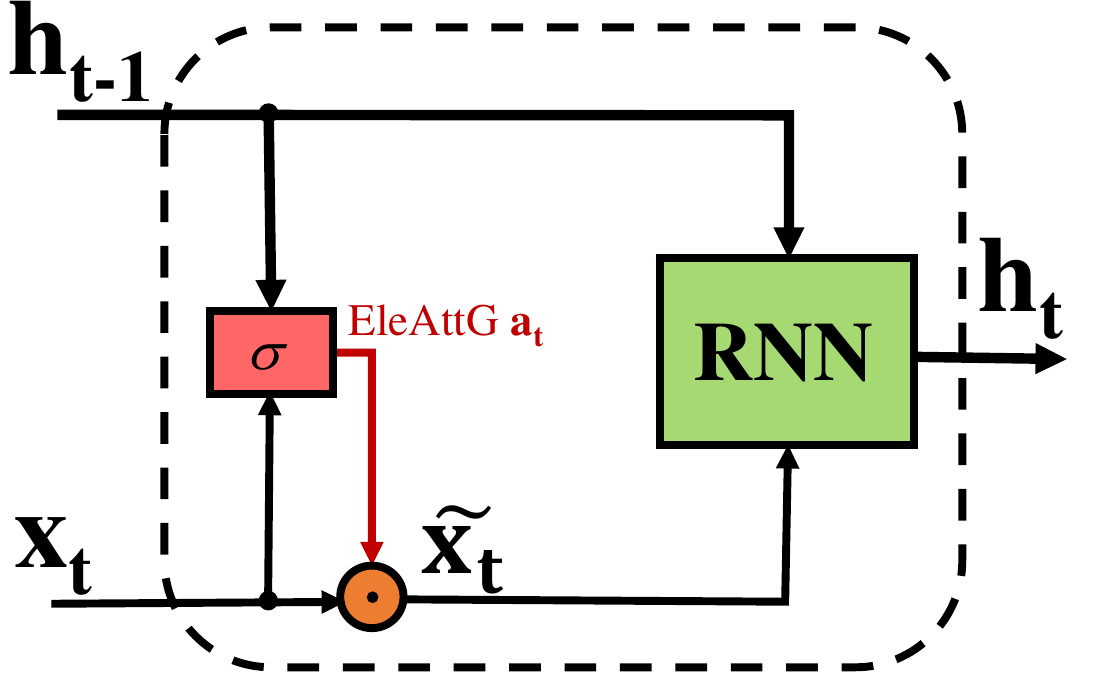}
		\caption{}
		\label{subfig:generalRNN}
	\end{subfigure}	
	\begin{subfigure}[t]{0.52\linewidth}
		\centering\includegraphics[width=\textwidth]{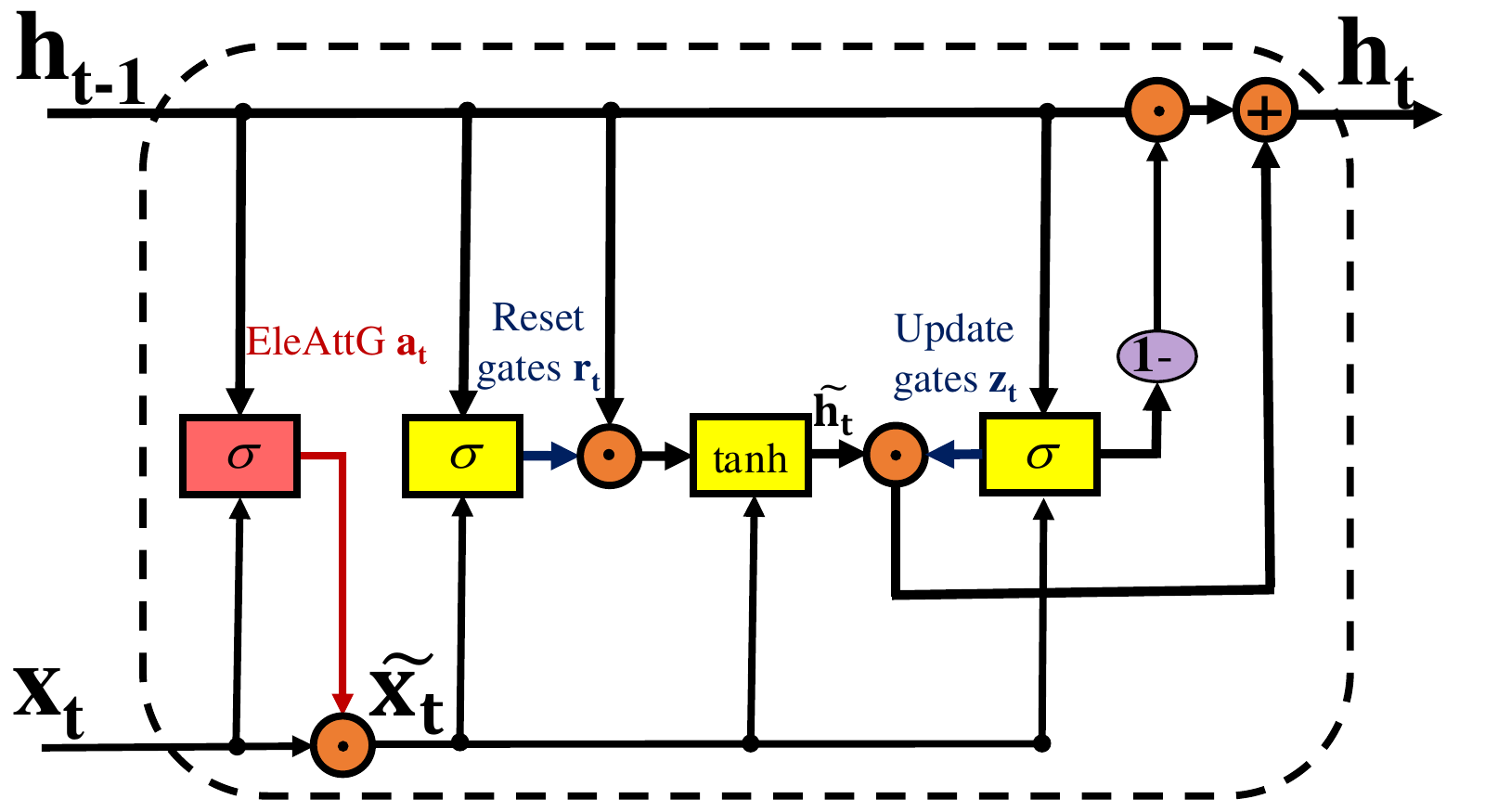}
		\caption{}			
		\label{subfig:EleGGRU}
	\end{subfigure}
	\caption[]{Illustration of \Outer-Attention Gate (EleAttG) (marked in red) for (a) a generic RNN block, where the RNN structure could be the standard RNN, LSTM, or GRU and (b) a GRU block which consists of a group of ({\it e.g.,} $N$) GRU neurons. In the diagram, each line carries a vector. The brown circles denote element-wise operation, {\it e.g.,} element-wise vector product or vector addition. The yellow boxes denote the units of the original GRU with the output dimension of $N$. The red box denotes the EleAttG with an output dimension of $D$, which is the same as the dimension of the input $\bf{x}_t$.}\label{fig:IAttRNN}
\end{figure}

The structure of recurrent neural networks facilitates the processing of sequential data. RNN neurons perform the same task at each step, with the output being dependent on the previous output, {\it i.e.}, {{some} historical information is memorized. Standard RNNs have difficulties in learning long-range dependencies due to the vanishing gradient problem \cite{jozefowicz2015empirical}. The LSTM \cite{hochreiter1997long} or GRU \cite{cho2014learning} architectures combat vanishing gradients through a gating mechanism. Gates provide a way to optionally let information through or stop softly, which balances the contributions of the information of the current time slot and historical information. There are some variants of RNNs with slightly different designs \cite{jozefowicz2015empirical}. {Note a gate applies a single scaling factor to control the flow of the embedded information (as a whole) of the input rather than imposing controls on each element of the input.} They are not designed to explore the potential different characteristics of the input elements.
	
	Attention mechanisms which selectively focus on different parts
	of the data have been demonstrated to be effective for many tasks \cite{luong2015effective,vaswani2017attention,xu2015show,li2017attentive,sharma2015actionattention,wang2016hierarchical}. These inspire us to develop an \Outer-Attention Gate (\EleAttGn) to augment the capability of RNN neurons. More specifically, for an RNN block, an EleAttG is designed to output an attention vector, with the same dimension as the input, which is then used to modulate the input elements. Note that similar to \cite{LSTMblog}, we use an RNN block to represent an ensemble of $N$ RNN neurons, which for example could be all the RNN neurons in an RNN layer. Fig.~\ref{fig:IAttRNN}~(a) illustrates the EleAttG within a generic RNN block. Fig.~\ref{fig:IAttRNN}~(b) shows a specific case when the RNN structure of GRU is used.  The input $\bf{x}_t$ is first modulated by the response of the EleAttG to output $\widetilde{\bf{x}_t}$ before other operations are applied to the RNN block. We refer to an RNN block equipped with an EleAttG as EleAtt-RNN block. Depending on the underlying RNN structure used ({\it e.g.}, standard RNN, LSTM, GRU), the newly developed EleAtt-RNN will also be denoted as EleAtt-sRNN, EleAtt-LSTM, or EleAtt-GRU. An RNN layer with such EleAttG can replace the original RNN layer and multiple EleAtt-RNN layers can be stacked.

	We demonstrate the effectiveness of the proposed EleAtt-RNN by applying it to action recognition. Specifically, for 3D skeleton-based human action recognition, we build our systems by stacking several EleAtt-RNN layers, using standard RNN, LSTM and GRU, respectively. EleAtt-RNNs consistently outperform the original RNNs for all the three types of RNNs. Our scheme based on EleAtt-GRU achieves state-of-the-art performance on three challenging datasets, {\it i.e.,} the NTU~\cite{Shahroudy_2016_CVPR}, N-UCLA~\cite{wang2014cross}, and SYSU~\cite{hu2015jointly} datasets. For RGB-based action recognition, we design our system by applying an EleAtt-GRU network to the sequence of frame-level CNN features. Experiments on both the JHMDB \cite{jhuang2013towards} and NTU \cite{Shahroudy_2016_CVPR} datasets show that adding {\EleAttGn}s brings significant gain.

	The proposed \EleAttG has the following merits. First, \EleAttG is capable of adaptively modulating the input at a fine granularity, paying different levels of attention to different elements of the input, resulting in faster convergence in training and higher performance. Second, the design is very simple. For an RNN layer, only one line of code needs to be added in implementation. Third, the EleAttG is general and can be added to any underlying RNN structure, {\it e.g.}, standard RNN, LSTM, GRU, and to any layer.      
	
	\section{Related work}
	\subsection{Recurrent Neural Networks}
	
	Recurrent neural networks have many different structures. In 1997, to address the vanishing gradient problem of standard RNN, Hochreiter {\it et al.} proposed LSTM, which introduces a memory cell that allows ``constant error carrousels" and multiplicative gate units that learn to open and close access to the constant error flow \cite{hochreiter1997long}. Gers {\it et al.} made improvement by adding the ``forget gate" that enables an LSTM cell to learn to reset itself (historical information) at appropriate times to prohibit the growth of the state indefinitely \cite{gers1999learning}. A variant of LSTM is the peephole LSTM, which allows the gates to access the cell \cite{gers2002learning}. GRU, which was proposed in 2014, is a simpler variant of LSTM. A GRU has a reset gate and an update gate which control the memory and the new input information. Between the LSTMs and GRUs, there is no clear winner \cite{chung2014empirical,jozefowicz2015empirical}. For LSTM, a differential gating scheme is proposed in \cite{veeriah2015differential} which leverages the derivative of the cell state to gate the information flow. Its effectiveness is demonstrated on action recognition. 
	
	In this work, we address the capability of RNNs from a new perspective. We propose a simple yet effective \Outer-Attention Gate which adaptively modulates the input elements to explore their different importances for an RNN block. 
	
	\subsection{Attention Mechanisms}
	Attention mechanisms which selectively focus on different parts of the data have been proven effective for many tasks such as machine translation \cite{luong2015effective,vaswani2017attention}, image caption \cite{xu2015show}, object detection \cite{li2017attentive}, and action recognition \cite{sharma2015actionattention,wang2016hierarchical}. 
	
	Luong {\it et al.} examine some simple attention mechanisms for neural machine translation. At each time step, the model infers the attention weights and uses them to average the embedding vectors of the source words \cite{luong2015effective}. For image caption, Xu {\it et al.} split an image into $L$ parts with each part described by a feature vector. To allow the decoder which is built by LSTM blocks to selectively focus on certain parts of the image, the weighted average of all the feature vectors using learned attention weights is fed to the LSTM network at every time step \cite{xu2015show}. A similar idea is used for RGB-based action recognition in \cite{sharma2015actionattention}. The above attention models focus on how to  average a set of feature vectors with suitable weights to generate a pooled vector of the same dimension as the input of RNN. They do not consider the fine-grained adjustment based on different levels of importance across the input dimensions. In addition, they address attention at the network level, but not RNN block level.
	
	For skeleton-based action recognition, a global context-aware attention is proposed to allocate different levels of importance to different joints of different input frames \cite{liu2017global}. Since the global information of a sequence is required to learn the attention, the system suffers from time delay. Song {\it et al.} propose a spatio-temporal attention model without requiring global information \cite{song2017end}. Before the main recognition network, a spatial attention subnetwork is added which modulates the skeleton input to selectively focus on discriminative joints at each time slot. However, their design is not general and has not been extended to higher RNN layers. In contrast, our proposed enhanced RNN, with EleAttG embedded as a fundamental unit of RNN block, is general, simple yet effective, which can be applied to any RNN block/layer. 
	
	\subsection {Action Recognition}
	\label{subsec:action}
	For action recognition, many studies focus on recognition from RGB videos \cite{wang2013action,simonyan2014two,donahue2015long,yue2015beyond,wang2016temporal}. In recent years, 3D skeleton based human action recognition has been extensively studied and has been attracting increasing attention, thanks to its high level representation \cite{han2017space}. Many traditional approaches focus on how to design efficient features to solve the problems of small inter-class variation, large view variations, and the modeling of complicated spatial and temporal evolution  \cite{wang2013action,wang2014cross,vemulapalli2014human,xia2012view,wang2012mining,wang2016graph}. 
	
	For 3D skeleton based human action recognition, RNN-based approaches have been attractive due to their powerful spatio-temporal dynamic modeling ability. Du {\it et al.} propose a hierarchical RNN model with the hierarchical body partitions as input to different RNN layers \cite{du2015hierarchical}. To exploit the co-occurrence of discriminative joints of skeletons, Zhu {\it et al.} propose a deep regularized LSTM networks with group sparse regularization \cite{zhu2016co}. Shahroudy {\it et al.} propose a part-aware LSTM network by separating the original LSTM cell into five sub-cells corresponding to five major groups of human body \cite{Shahroudy_2016_CVPR}. Liu {\it et al.} propose a spatio-temporal LSTM structure to explore the contextual dependency of joints in spatio-temporal domains \cite{liu2016spatio}. Li {\it et al.} propose an RNN tree network with a hierarchical structure which classifies the action classes that are easier to distinguish at the lower layers and the action classes that are harder to distinguish at higher layers \cite{li2017adaptive}. To address the large view variation of the captured data, Zhang {\it et al.} propose a view adaptive subnetwork which automatically selects the best observation viewpoints within an end-to-end network for recognition \cite{zhang2017view}. 
	
	For RGB-based action recognition, to exploit the spatial correlations, convolutional neural networks are usually used to learn the features \cite{simonyan2014two,wang2016temporal,yue2015beyond,donahue2015long}. Some approaches explore the temporal dynamics of the sequential frames by simply averaging/multiplying the scores/features of the frames for fusion \cite{simonyan2014two,wang2016temporal,diba2017deeptemporal}. Some other approaches leverage RNNs to model temporal correlations, with frame-wise CNN features as input at every time slot \cite{yue2015beyond,donahue2015long}. 
	
	Our proposed \EleAttG is a fundamental unit that aims to enhance the capability of an RNN block. We will demonstrate its effectiveness in both 3D skeleton based action recognition and RGB-based action recognition. 
	
	\section{Overview of Standard RNN, LSTM, and GRU}
	\label{sec:RNN} 
	Recurrent neural networks are capable of modeling temporal dynamics of a time sequence. They have a ``memory" which captures historical information accumulated from previous time steps. To better understand the proposed EleAttG and its generalization capability, we briefly review the popular RNN structures, {\it i.e.,} standard RNN, Long Short Term Memory (LSTM) \cite{hochreiter1997long}, and Gated Recurrent Unit (GRU) \cite{cho2014learning}.
	
	For a standard RNN layer, the output response $\bf{h}_t$ at time $t$ is calculated based on the input $\bf{x}_t$ to this layer and the output $\bf{h}_{t-1}$ from the previous time slot

	\begin{equation}
	\label{equ:rnn}
	{{\bf{h}}_{t}} = \tanh \left( {{\bf{W}}_{xh}}{{\bf{x}}_{t}} + {{\bf{W}}_{hh}}{{\bf{h}}_{t-1} + {\bf{b}}_h} \right),
	\end{equation}
	where ${\bf{W}}_{\alpha\beta}$ denotes the matrix of weights between $\alpha$ and $\beta$, $\bf{b}_h$ is the bias vector.

	The standard RNN suffers from the gradient vanishing problem due to insufficient, decaying error back flow \cite{hochreiter1997long}. LSTM allevates this problem by enforcing constant error flow through ``constant error carrousels" within the cell unit ${c}_t$. The input gate ${i}_t$, forget gate ${f}_t$ and output gate ${o}_t$ learn to open and close access to the constant error flow. For an LSTM layer, the recursive computations of activations of the units are
	\begin{eqnarray}
	\label{equ:lstm}
	\begin{aligned}
	\!&{\bf{i}}_t = \sigma \left( {{\bf{W}}_{xi}}{{\bf{x}}_{t}} + {{\bf{W}}_{hi}}{\bf{h}}_{t-1} + {\bf{b}}_i \right), \\ 
	\!&{\bf{f}}_t = \sigma \left( {{\bf{W}}_{xf}}{{\bf{x}}_{t}} + {{\bf{W}}_{hf}}{\bf{h}}_{t-1} + {\bf{b}}_f \right), \\
	\!&{\bf{c}}_t = {\bf{f}}_t\!\odot {\bf{c}}_{t-1}\!+{\bf{i}}_t \odot \tanh\! \left( {{\bf{W}}_{xc}}{{\bf{x}}_{t}}\! +\! {{\bf{W}}_{hc}}{\bf{h}}_{t-1} \!+\! {\bf{b}}_c \right), \\ 
	\!&{\bf{o}}_t = \sigma \left( {{\bf{W}}_{xo}}{{\bf{x}}_{t}} + {{\bf{W}}_{ho}}{\bf{h}}_{t-1} + {\bf{b}}_o \right), \\
	\!&{\bf{h}}_t = {\bf{o}}_t \odot \tanh \left( {\bf{c}}_t \right),
	\end{aligned}
	\end{eqnarray}
	where $\odot$ denotes an element-wise product. Note that $\bf{i}_t$ is a vector denoting the responses of a set of input gates of all the LSTM neurons in the layer.
	
	GRU is an architecture that is similar to but much simpler than that of LSTM. A GRU has two gates, reset gate $r_t$ and update gate $z_t$. When the response of the reset gate is close to 0, the hidden state $h_t'$ is forced to ignore the previous hidden state and reset with the current input only. The update gate controls how much information from the previous hidden state will be carried over to the current hidden state $h_t$. The hidden state acts in a way similar to the memory cell in LSTM. For a GRU layer, the recursive computations of activations of the units are
	\begin{equation}
	\label{equ:gru}
	\begin{aligned}
	\!&\mathbf{r}_t = \sigma \left( {\mathbf{W}}_{xr} \mathbf{x}_t + {\mathbf{W}}_{hr} \mathbf{h}_{t-1} + \mathbf{b}_r \right),\\
	\!&\mathbf{z}_t = \sigma \left( {\mathbf{W}}_{xz} \mathbf{x}_t + {\mathbf{W}}_{hz} \mathbf{h}_{t-1} + \mathbf{b}_z \right),\\
	\!&\mathbf{h}_t' = \tanh \left( {\mathbf{W}}_{xh} \mathbf{x}_t + {\mathbf{W}}_{hh} (\mathbf{r}_t \odot \mathbf{h}_{t-1}) + \mathbf{b}_h \right),\\
	\!&\mathbf{h}_t =  {\mathbf{z_t}} \odot {\mathbf{h}_{t-1}}  + ({\mathbf{1 - z_t}}) \odot {\mathbf{h}_t'}.\\
	\end{aligned}
	\end{equation}

	For all the above designs, we note that the gates can control the information flow. However, the controlling of the flow takes the input $\bf{x}_t$ as a whole without adaptively treating different elements of the input differently. 
	
	\section{\Outer-Attention Gate for an RNN Block}
	\label{sec:EleAtt-RNN}

	For an RNN block, we propose an \Outer-Attention Gate (\EleAttGn) to enable the RNN neurons to have the attentiveness capability. The response of an \EleAttG is a vector $\bf{a}_t$ with the same dimension as the input $\bf{x}_t$ of the RNNs, which is calculated as 
	\begin{eqnarray}
	\label{equ:agru}
	\begin{aligned}
	\!&\bf{a}_t =\phi \left({\bf{W}_{xa}}{\bf{x}_t} + {\bf{W}_{ha} {\bf{h}_{t-1}}} + {\bf{b}_a}  \right),
	\end{aligned}
	\end{eqnarray}
	where $\phi$ denotes the activation function of Sigmoid, {\it i.e.,} $\phi(s) = 1/(1+e^{-s}$). $\bf{W}_{\alpha\beta}$ denotes the matrix of weights between $\alpha$ and $\beta$, and $\bf{b}_a$ denotes the bias vector. The current input $\bf{x}_t$ and the hidden states $\bf{h}_{t-1}$ are used to determine the levels of importance of each element of the input $\bf{x}_t$.
	
	The attention response modulates the input to have an updated input $\widetilde{\bf{x}_t}$ as 
	\begin{eqnarray}
	\label{equ:updatedx}
	\begin{aligned}
	\!&\widetilde{\bf{x}_t}= \bf{a}_t \odot {\bf{x}_t}.
	\end{aligned}
	\end{eqnarray}
	The recursive computations of activations of the other units in the RNN block are then based on the updated input $\widetilde{\bf{x}_t}$, instead of the original input $\bf{x}_t$, as illustrated in Fig. \ref{fig:IAttRNN}.
	
	For a standard RNN block with EleAttG (denoted as EleAtt-sRNN), the output responses $\bf{h}_t$ at time $t$ are calculated as
	\begin{equation}
	\label{equ:EleAtt-rnn}
	{{\bf{h}}_{t}} = \tanh \left( {{\bf{W}}_{xh}}\widetilde{\bf{x}_t} + {{\bf{W}}_{hh}}{{\bf{h}}_{t-1} + {\bf{b}}_h} \right).
	\end{equation}
	
	Similarly, for an EleAtt-GRU block, the recursive computations of activations of the units are
	\begin{equation}
	\label{equ:EleAtt-gru}
	\begin{aligned}
	\!&\mathbf{r}_t = \sigma \left( {\mathbf{W}}_{xr} \widetilde{\bf{x}_t} + {\mathbf{W}}_{hr} \mathbf{h}_{t-1} + \mathbf{b}_r \right),\\
	\!&\mathbf{z}_t = \sigma \left( {\mathbf{W}}_{xz} \widetilde{\bf{x}_t} + {\mathbf{W}}_{hz} \mathbf{h}_{t-1} + \mathbf{b}_z \right),\\
	\!&\mathbf{h}_t' = \tanh \left( {\mathbf{W}}_{xh} \widetilde{\bf{x}_t} + {\mathbf{W}}_{hh} (\mathbf{r}_t \odot \mathbf{h}_{t-1}) + \mathbf{b}_h \right),\\
	\!&\mathbf{h}_t =  {\mathbf{z_t}} \odot {\mathbf{h}_{t-1}}  + ({\mathbf{1 - z_t}}) \odot {\mathbf{h}_t'}.\\
	\end{aligned}
	\end{equation}
	
	The computations for an EleAtt-LSTM block can be obtained similarly.
	
	Most attention designs use Softmax as the activation function such that the sum of the attention values is 1 \cite{luong2015effective,vaswani2017attention,xu2015show,li2017attentive,sharma2015actionattention,wang2016hierarchical,song2017end}. In our design, we relax this sum-to-1 constraint by using the Sigmoid activation function, with response values ranging from 0 to 1. If the sum-to-1 constraint is not relaxed, the attention responses of the $k^{th}$ element will be affected by the changes of other elements' response values even when the levels of importance of this element are the same over consecutive time slots. 

	Note that in our design, an \EleAttG is shared by all neurons in an RNN block/layer (see (\ref{equ:updatedx}) and (\ref{equ:EleAtt-rnn}) for the standard RNN block, (\ref{equ:updatedx}) and (\ref{equ:EleAtt-gru}) for the GRU block). Theoretically, each RNN neuron (instead of block) can have its own attention gate at the cost of increased computation complexity and a larger number of parameters. We focus on the shared design in this work.}

\section{Experiments}
We perform comprehensive studies to evaluate the effectiveness of our proposed \EARNN~with \EleAttG by applying it to action recognition from 3D skeleton data and RGB video, respectively. 

To demonstrate the generalization capability of \EleAttGn, we add \EleAttG to the standard RNN, LSTM, and GRU structures, respectively. 

For 3D skeleton based action recognition, we use three challenging datasets, {\it{i.e.},} the NTU RGB+D dataset (NTU) \cite{Shahroudy_2016_CVPR}, the Northwestern-UCLA dataset (N-UCLA) \cite{wang2014cross}, and the SYSU Human-Object Interaction dataset (SYSU)\cite{hu2015jointly}. The NTU is currently the largest dataset with diverse subjects, various viewpoints and small inter-class differences. Therefore, in-depth analyses are performed on the NTU dataset. For RGB-based action recognition, we take the CNN features extracted from existing, pre-trained models without finetuning on our datasets as the input to the RNN based recognition networks and evaluate the effectiveness of \EleAttG on the NTU and the JHMDB datasets \cite{jhuang2013towards}. We conduct most of our experiments based on GRU here, as it has simpler structure than LSTM and better performance than standard RNN.

\subsection{Datasets}
\label{datasets}
\textbf{NTU RGB+D Dataset (NTU)~\cite{Shahroudy_2016_CVPR}.} NTU is currently the largest RGB+D+Skeleton dataset for action recognition, including 56880 videos of in total more than 4 million frames. There are 60 action classes performed by different subjects. Each subject has 25 body joints and each joint has 3D coordinates. Three cameras placed in different positions are used to capture the data at the same time. We follow the standard protocols proposed in \cite{Shahroudy_2016_CVPR} including the Cross Subject (CS) and Cross View (CV) settings. For the CS setting, 40 subjects are equally split into training and testing groups. For the CV setting, the samples of cameras 2 and 3 are used for training while those of camera 1 are for testing.

\textbf{Northwestern-UCLA dataset (N-UCLA)~\cite{wang2014cross}.} N-UCLA is a small RGB+D+Skeleton dataset including 1494 sequences which records 10 actions performed by 10 subjects. 20 joints with 3D coordinates are provided in this dataset. Following \cite{wang2014cross}, we
use samples from the first two cameras as training data, and the
samples from the third camera as testing data. 

\textbf{SYSU Human-Object Interaction dataset (SYSU)~\cite{hu2015jointly}.} SYSU is a small RGB+D+Skeleton dataset, including 480 sequences performed by 40 different subjects. It contains 12 actions. A subject has 20 joints with 3D coordinates. We follow the standard protocols proposed in \cite{hu2015jointly} for evaluation. They include two settings. For the Cross Subject (CS) setting, half of the subjects are used for training and the others for testing. For the Same Subject (SS) setting, half of the sequences of each subject are used for training and others for testing. The average performance of 30-fold cross validation is reported.

\textbf{JHMDB dataset (JHMDB)~\cite{jhuang2013towards}.} JHMDB is an RGB-based dataset which has 928 RGB videos with each video containing about 15-40 frames. It contains 21 actions performed by different actors. This dataset is challenging where the videos are collected on the Internet which also includes outdoor activities.

\subsection{Implementation Details}
\label{subsec:implement}

We perform our experiments on the deep learning platform of Keras \cite{chollet2015keras} with Theano \cite{al2016theano} as the backend. For the RNN networks, Dropout \cite{srivastava2014dropout} with the probability of 0.5 is used to alleviate overfitting. Gradient clipping similar to \cite{sutskever2014sequence} is used by constraining the maximum amplitude of the gradient to 1. Adam \cite{kingma2014adam} is used to train the networks from end-to-end. The initial learning rate is set to 0.005 for 3D skeleton-based action recognition and 0.001 for RGB-based action recognition. During training, the learning rate will be reduced by a factor of 10 when the training accuracy does not increase. We use cross-entropy as the loss function. 

For 3D skeleton-based action recognition, similar to the classification network design in \cite{zhang2017view}, we build our systems by stacking three RNN layers with {\EleAttGn}s and one fully connected (FC) layer for classification. We use 100 RNN neurons in each layer. Considering the large difference on the sizes of the datasets, we set the batch size for the NTU, N-UCLA, and SYSU datasets to 256, 32, and 32, respectively. We use the sequence-level pre-processing method in \cite{zhang2017view} by setting the body center in the first frame as the coordinate origin to make the system invariant to the initial position of human body. To improve the robustness to view variations at the sequence level, we can perform data augmentation by randomly rotating the skeleton around the X, Y and Z axes by various degrees ranging from -35 to 35 during training. For the N-UCLA and SYSU datasets, we use the RNN models pre-trained on the NTU dataset to initialize the baseline schemes and the proposed schemes.

\begin{figure}[t] 
	\setlength{\belowcaptionskip}{-10pt}
	\centering
	\begin{subfigure}[t]{0.45\linewidth}
		\centering\includegraphics[width=\textwidth]{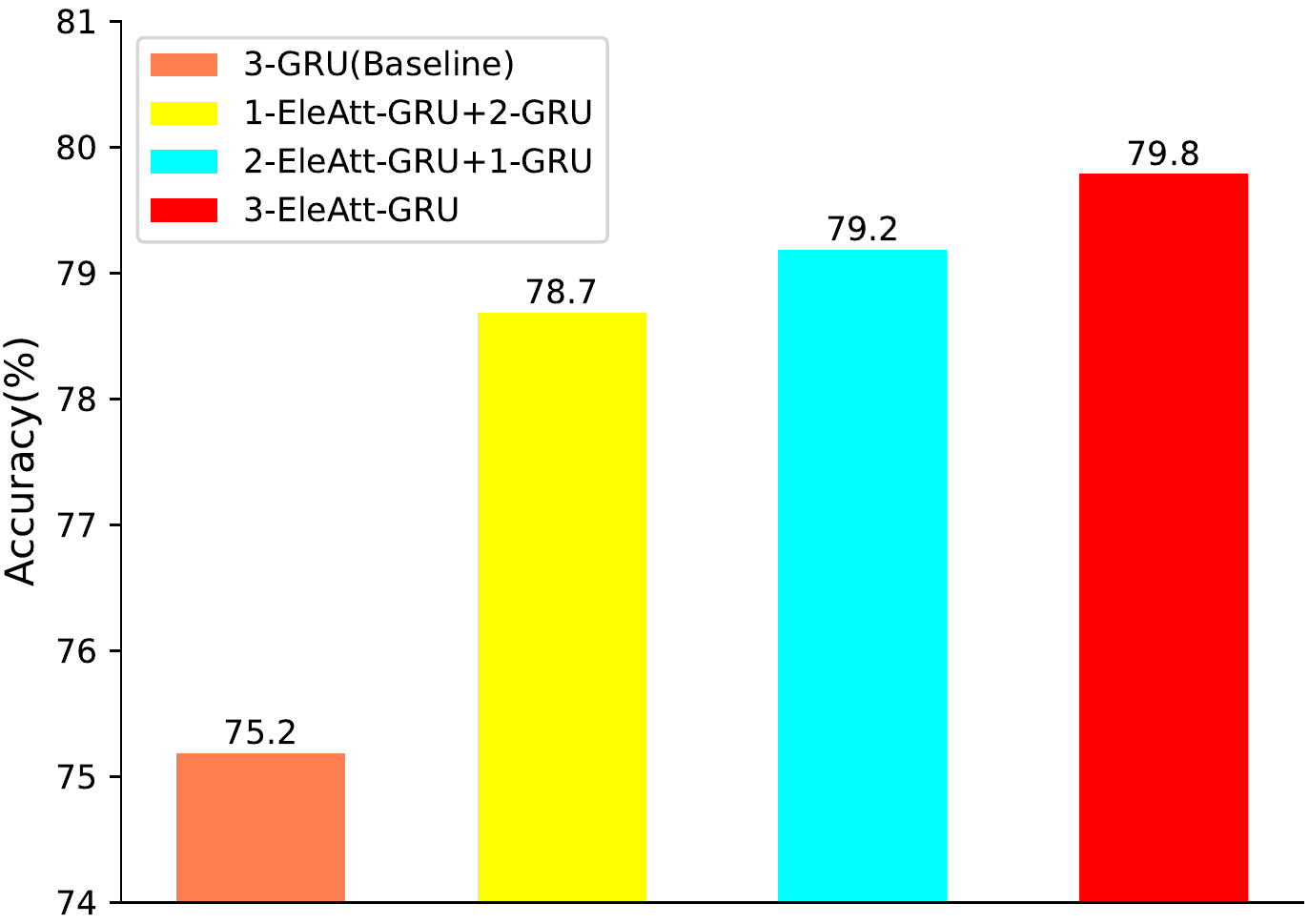}
		\caption{NTU-CS}
		\label{subfig:CS}
	\end{subfigure}	
	~~
	\begin{subfigure}[t]{0.45\linewidth}
		\centering\includegraphics[width=\textwidth]{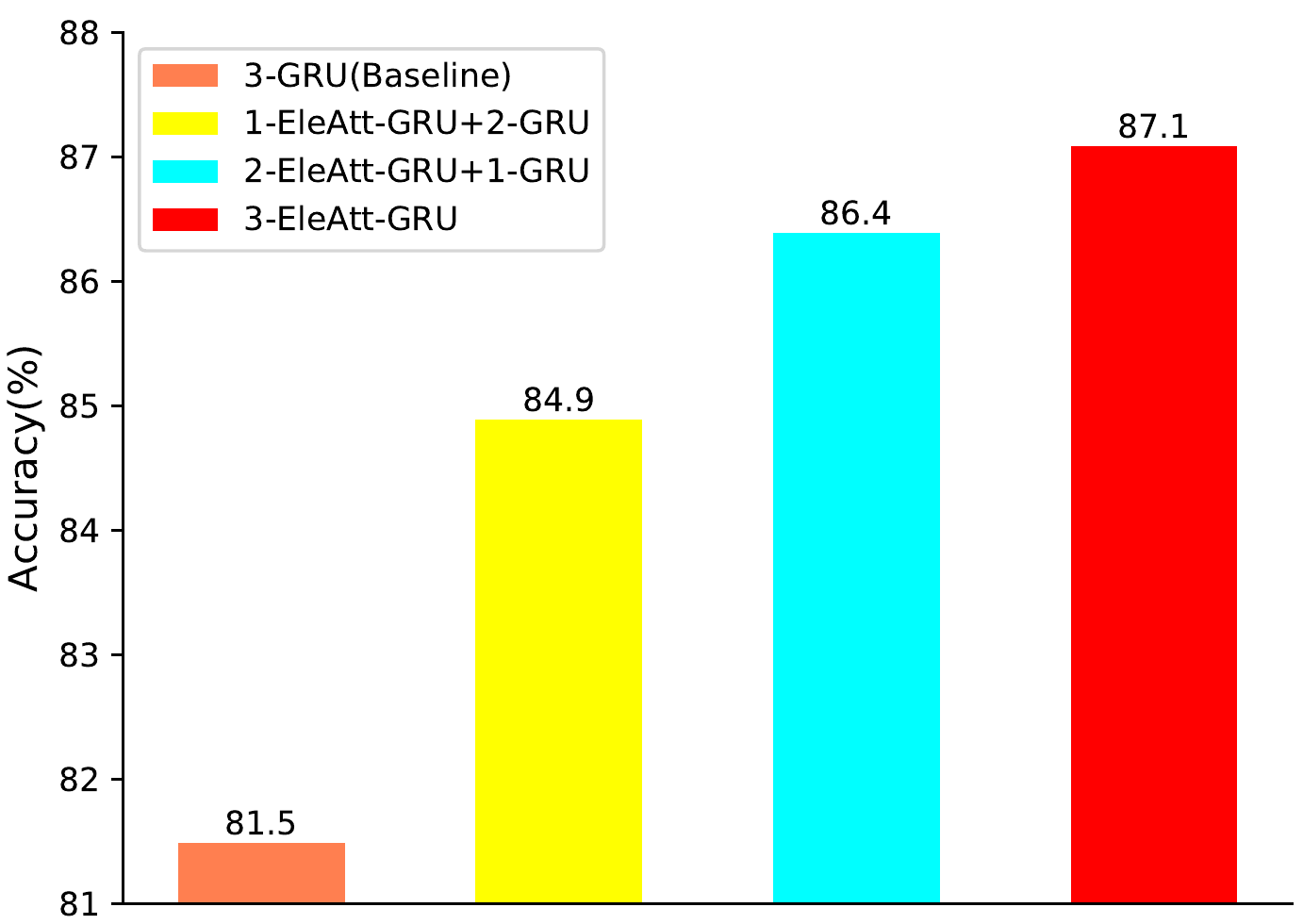}
		\caption{NTU-CV}			
		\label{subfig:CV}
	\end{subfigure}
	\caption{Effectiveness of proposed {\EleAttGn}s on the three layered GRU network for 3D skeleton based human action recognition on the NTU dataset. ``$m$-EleAtt-GRU+$n$-GRU" denotes that the first $m$ layers are EleAtt-GRU layers and the remaining $n$ layers are the original GRU layers.}\label{fig:layer}
\end{figure}

For RGB-based action recognition, we feed an RNN network with the features to further explore temporal dynamics. Since our purpose is to evaluate whether the proposed \EleAttG can generally improve recognition accuracy, we extract CNN features using some available pre-trained models without finetuning for the specific dataset or task. For the JHMDB dataset, we use the TSN model from \cite{wang2016temporal,TSNModel} which was trained on the HMDB dataset \cite{kuehne2011hmdb} to extract a 1024 dimensional feature for each frame. For the NTU dataset which has more videos, we take the ResNet50 model \cite{he2016deep,ResNet50Model} which has been pre-trained on ImageNet as our feature extractor (2048 dimensional feature for each frame) considering the ResNet50 model is much faster than the TSN model. The implementation details of the RNN networks are similar to that discussed above. For the NTU dataset, we stack three EleAtt-GRU layers, with each layer consisting of 512 GRU neurons. For the JHMDB dataset, we use only one GRU layer (512 GRU neurons) with \EleAttG to avoid overfitting, considering that the number of video samples is much smaller than that of the NTU dataset. The batch size is set to 256 for the NTU dataset and 32 for the JHMDB dataset.

\subsection{Effectiveness of \Outer-Attention-Gates} 



\noindent\textbf{Effectiveness on GRU network.} Fig. \ref{fig:layer} shows the effectiveness of {\EleAttGn} on the GRU network. Our final scheme with three EleAtt-GRU layers (``3-EleAtt-GRU") outperforms the baseline scheme ``3-GRU(Baseline)" significantly, by {\bf{4.6\%}} and {\bf{5.6\%}} for the CS and CV settings, respectively. The performance increases as more GRU layers are replaced by the EleAtt-GRU layers.

\noindent\textbf{Generalization to other input signals.} The proposed RNN block with {\EleAttGn} is generic and can be applied to different types of source data. To demonstrate this, we use CNN features extracted from RGB frames as the input  of the RNNs for RGB based action recognition. Table \ref{tab:rgb} shows the performance comparisons on the NTU and JHMDB dataset respectively. The implementation details have been described in Section \ref{subsec:implement}. We can see that the ``EleAtt-GRU" outperform the ``Baseline-GRU" by about 2-4\% on the NTU dataset, and 2\% on the JHMDB dataset. Note that the performance is not optimized since we have not used the fine-tuned CNN model on this dataset for this task. 

\setlength{\tabcolsep}{7pt}
\begin{table}[t]
	\centering
	\caption{Effectiveness of proposed {\EleAttGn}s in the GRU network for RGB-based action recognition on the NTU and JHMDB datasets.}
	\label{tab:rgb}
	\begin{tabular}{ccccccc}
		\toprule
		\multirow{2}{*}{Dataset} & \multicolumn{2}{c}{NTU} & \multicolumn{4}{c}{JHMDB}        \\
		\cmidrule(lr){2-3}
		\cmidrule(lr){4-7}
		& CS         & CV         & Split1 & Split2 & Split3 & Average  \\
		\midrule
		Baseline-GRU                & 61.3      & 66.8      & 60.6  & 59.2  & 62.9  & 60.9 \\
		EleAtt-GRU                    & 63.3      & 70.6      & 64.5  & 59.2  & 65.0  & 62.9 \\
		\bottomrule
	\end{tabular}
\end{table}

\setlength{\tabcolsep}{7pt}
\begin{table}[t] 
	\centering
	\caption{Effectiveness of {\EleAttGn}s on three types of RNN structures on the NTU dataset. ``EleAtt-$X$" denotes the scheme with {\EleAttGn}s based on the RNN structure of $X$.}
	\label{tab:Extend}
	\begin{tabular}{cccc}
		\toprule
		RNN structure                    & Scheme     & CS & CV \\
		\midrule
		\multirow{2}{*}{Standard RNN} & Baseline(1-sRNN)   & 51.6    & 57.6 \\
		& EleAtt-sRNN & \textbf{61.6}     & \textbf{67.2}  \\
		\midrule
		\multirow{2}{*}{LSTM}      & Baseline(3-LSTM)   & 77.2     & 83.0  \\
		& EleAtt-LSTM     & \textbf{78.4}     & \textbf{85.0}  \\
		\midrule
		\multirow{2}{*}{GRU}       & Baseline(3-GRU)   & 75.2     & 81.5  \\
		& EleAtt-GRU       & \textbf{79.8}     & \textbf{87.1} 							\\
		\bottomrule	
	\end{tabular}
\end{table}

\setlength{\tabcolsep}{7pt}
\begin{table}[t] 
	\centering
	\caption{Performance comparisons on the NTU dataset in accuracy (\%).}
	\label{tab:ntu}
	\begin{tabular}{ccc}
		\toprule
		{Method}                                           & CS & CV \\
		\midrule
		{Skeleton Quads \cite{evangelidis2014skeletal}}  & 38.6     & 41.4  \\
		{Lie Group \cite{vemulapalli2014human}}          & 50.1     & 52.8  \\
		{Dynamic Skeletons  \cite{hu2015jointly}}        & 60.2    & 65.2  \\
		{HBRNN-L  \cite{du2015hierarchical}}             & 59.1     & 64.0  \\
		{Part-aware LSTM   \cite{Shahroudy_2016_CVPR}} & 62.9     & 70.3  \\
		{ST-LSTM + Trust Gate \cite{liu2016spatio}}      & 69.2     & 77.7  \\
		{STA-LSTM \cite{song2017end}}                    & 73.4     & 81.2  \\
		{GCA-LSTM \cite{liu2017global}}                  & 74.4     & 82.8  \\
		{URNN-2L-T \cite{li2017adaptive}}                & 74.6     & 83.2  \\
		{Clips+CNN+MTLN \cite{ke2017new}}                & 79.6     & 84.8  \\
		{VA-LSTM \cite{zhang2017view}}                   & 79.4     & 87.2  \\
		\midrule
		Baseline-GRU  & 75.2 & 81.5 \\
		EleAtt-GRU & 79.8  & 87.1 \\
		EleAtt-GRU(aug.)     & \textbf{80.7}    & \textbf{88.4} \\
		\bottomrule
	\end{tabular}
\end{table}

\noindent\textbf{Generalization on various RNN structures.} The proposed \EleAttG is generic and can be applied to various RNN structures. We evaluate the effects of {\EleAttGn}s on three representative RNN structures, {\it i.e.,} the standard RNN (sRNN), LSTM, GRU respectively and show the results in Table \ref{tab:Extend}. Compared with LSTM and GRU, the standard RNN neurons do not have the gating designs which control the contributions of the current input to the network. The \EleAttG can element-wise control the contribution of the current input, which remedies the lack of gate designs to some extent. The gate designs in LSTM and GRU can only control the information flow input-wise. In contrast, the proposed {\EleAttGn}s are capable of controlling the input element-wise, adding the attentiveness capability to RNNs. We can see that the adding of {\EleAttGn}s  enhances performance significantly. Note that for sRNN, we build both the Baseline(1-sRNN) and our scheme using one sRNN layer rather than three as those for LSTM and GRU, in considering that the three layered sRNN baseline converges to a poorer performance, i.e., 33.6\% and 42.8\% for the CS and CV settings, which may be caused by the gradient vanishing of sRNN.

\noindent\textbf{Comparisons with state-of-the-arts on skeleton based action recognition.}
Table \ref{tab:ntu}, \ref{tab:N-UCLA} and \ref{tab:SYSU} show the performance comparisons with state-of-the-art approaches for the NTU, N-UCLA and SYSU datasets, respectively. ``Baseline-GRU" denotes our baseline scheme which is built by stacking three GRU layers while ``EleAtt-GRU" denotes our proposed scheme which replaces the GRU layers by the proposed GRU layers with {\EleAttGn}s. Implementation details can be found in Section \ref{subsec:implement}.  ``EleAtt-GRU(aug.)" denotes that data argumentation by rotating skeleton sequences is performed during training. We achieve the best performances in comparison with other state-of-the-art approaches on all the three datasets. Our scheme ``EleAtt-GRU" achieves significant gains over the baseline scheme ``Baseline-GRU", of 4.6-5.6\%, 4.7\%, and 2.4-2.8\% on the NTU, N-UCLA, and SYSU datasets, respectively.

\begin{figure}[t] 
	\centering
	\includegraphics[width=0.85\linewidth]{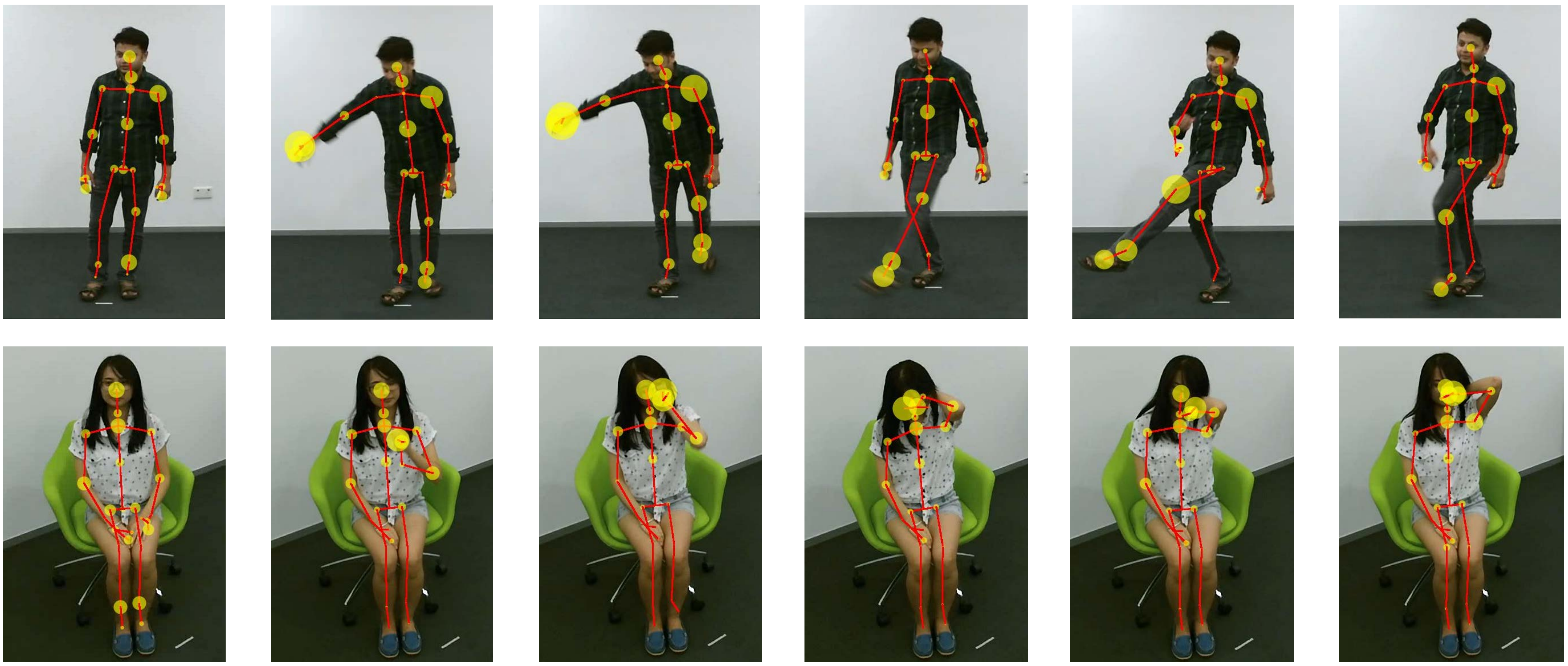}
	\caption{Visualization based on the attention responses of the first GRU layer for the actions of \emph{kicking} and \emph{touching neck}. For each joint, the size of the yellow circle indicates the learned level of importance.}	
	\label{fig:vis}
\end{figure}

\noindent\textbf{Visualization of the responses of \EleAttGn.} To better understand the learned element-wise attention, we observe the responses of the \EleAttG in the first GRU layer for the skeleton based action recognition. In the first layer, the input (with dimension of $3\times J$) at a time slot corresponds to the $J$ joints with each joint represented by the $X$, $Y$, and $Z$ coordinate values. The physical meaning of the attention responses is clear. However, in a higher layer, the \EleAttG modulates the input features on each element which is more difficult to interpret and visualize. Thus, we perform visualization based on the attention responses of the first GRU layer in Fig.~\ref{fig:vis} for the actions of \emph{kicking} and \emph{touching the neck}. 
Actually, the absolute response values cannot represent the relative importances across dimensions very well. The statistical energies of the different elements of the original input are different. For example, the foot joint which is in general far away from the body center has a higher average energy than that of the body center joint. We can imagine that there is a static modulation $\overline{a_i}$ on the $i^{th}$ element of the input, which can be calculated by the average energy before and after the modulation. For the $i^{th}$ element of an sample $j$ with attention value $a_{i,j}$, we use the relative response value $\widehat{a_{i,j}} = a_{i,j}/\overline{a_i}$ for visualization to better reflect the importances among joints. Note that the sum of the relative responses for the $X$, $Y$, and $Z$ of a joint is utilized for visualization. For the action of \emph{touching neck}  which is highly concerned with the joints on the arms and heads, the relative attention on those joints are larger. For \emph{kicking}, the relative attention on the legs is large. These are consistent with a human's perception.  

\setlength{\tabcolsep}{7pt}
\begin{table}[!t]
	\parbox{.44\linewidth}{
		\centering
		\caption{Performance comparisons on the N-UCLA dataset in acc. (\%).}
		\begin{tabular}{ccc}
			\toprule
			Method & accuracy \\ 
			\midrule
			HOJ3D \cite{xia2012view}          & 54.5      \\ 
			AE  \cite{wang2014learning}            & 76.0    \\ 
			VA-LSTM \cite{zhang2017view}            & 70.7 \\			
			HBRNN-L \cite{du2015hierarchical}        & 78.5     \\ 
			\midrule
			Baseline-GRU & 84.3 \\ 
			EleAtt-GRU & 89.0 \\ 
			EleAtt-GRU(aug.) & \bf{90.7}\\ 
			\bottomrule     
			\label{tab:N-UCLA}
		\end{tabular}
	}
	\hfill
	\parbox{.5\linewidth}{
		\centering
		\caption{Performance comparisons on the SYSU dataset in acc. (\%).}
		\begin{tabular}{ccc}
			\toprule
			Method & CS  & SS \\
			\midrule
			LAFF \cite{hu2016real} & 54.2 & - \\
			DS \cite{hu2015jointly} & 75.5 & 76.9 \\	
			ST-LSTM\cite{liu2016spatio} & 76.5 & - \\
			VA-LSTM \cite{zhang2017view} & 76.9 & 77.5 \\
			\midrule
			Baseline-GRU  &  82.1  & 82.1 \\
			EleAtt-GRU & 84.9 & 84.5 \\
			EleAtt-GRU(aug.) & \bf{85.7} & \bf{85.7} \\
			\bottomrule      
			\label{tab:SYSU}
		\end{tabular}
	}
\end{table}

\begin{figure}[!ttp]
	\centering
	\begin{subfigure}[t]{0.45\linewidth}
		\centering\includegraphics[width=\textwidth]{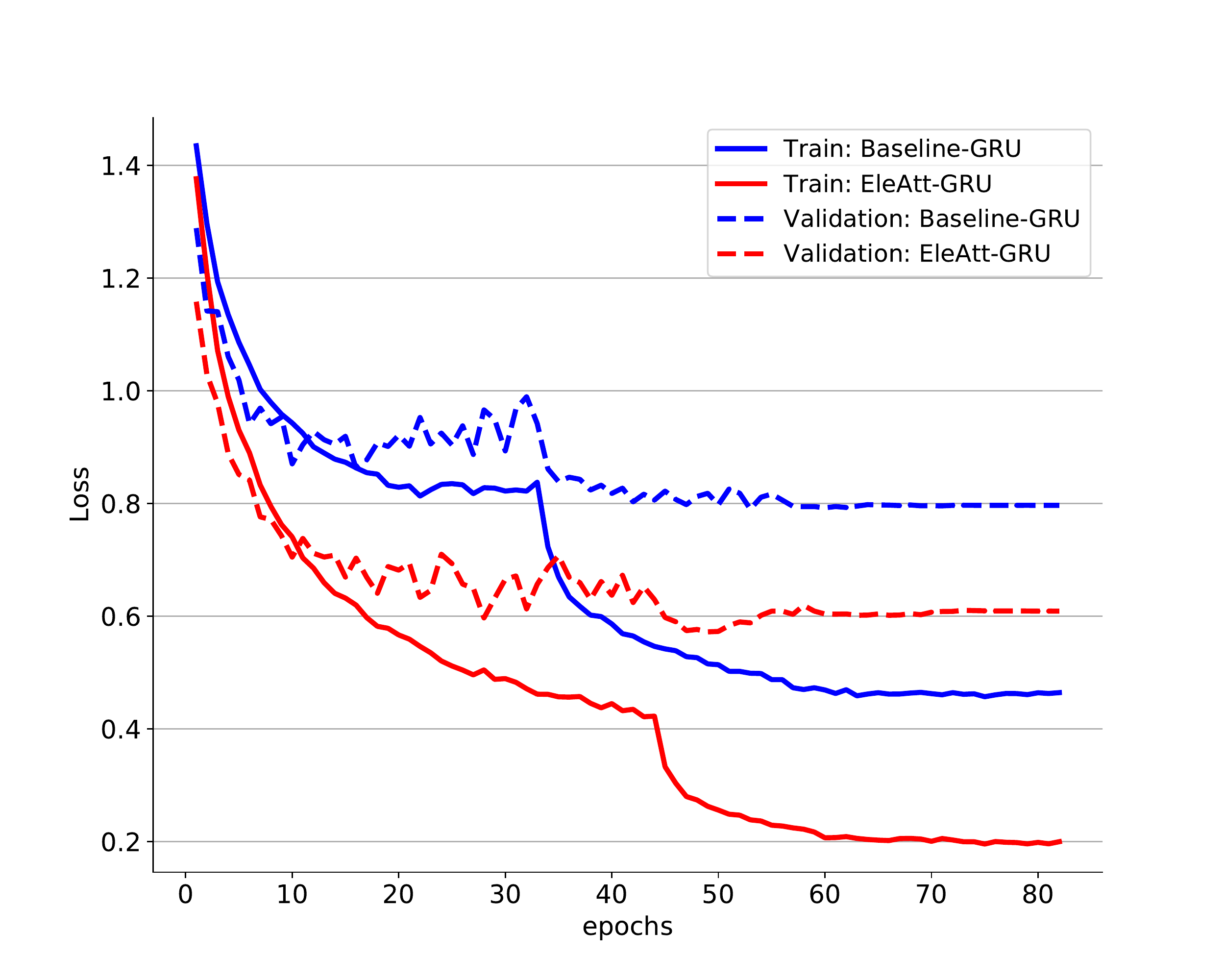}
		\caption{NTU-CS}
		\label{subfig:CS}
	\end{subfigure}	
	\begin{subfigure}[t]{0.45\linewidth}
		\centering\includegraphics[width=\textwidth]{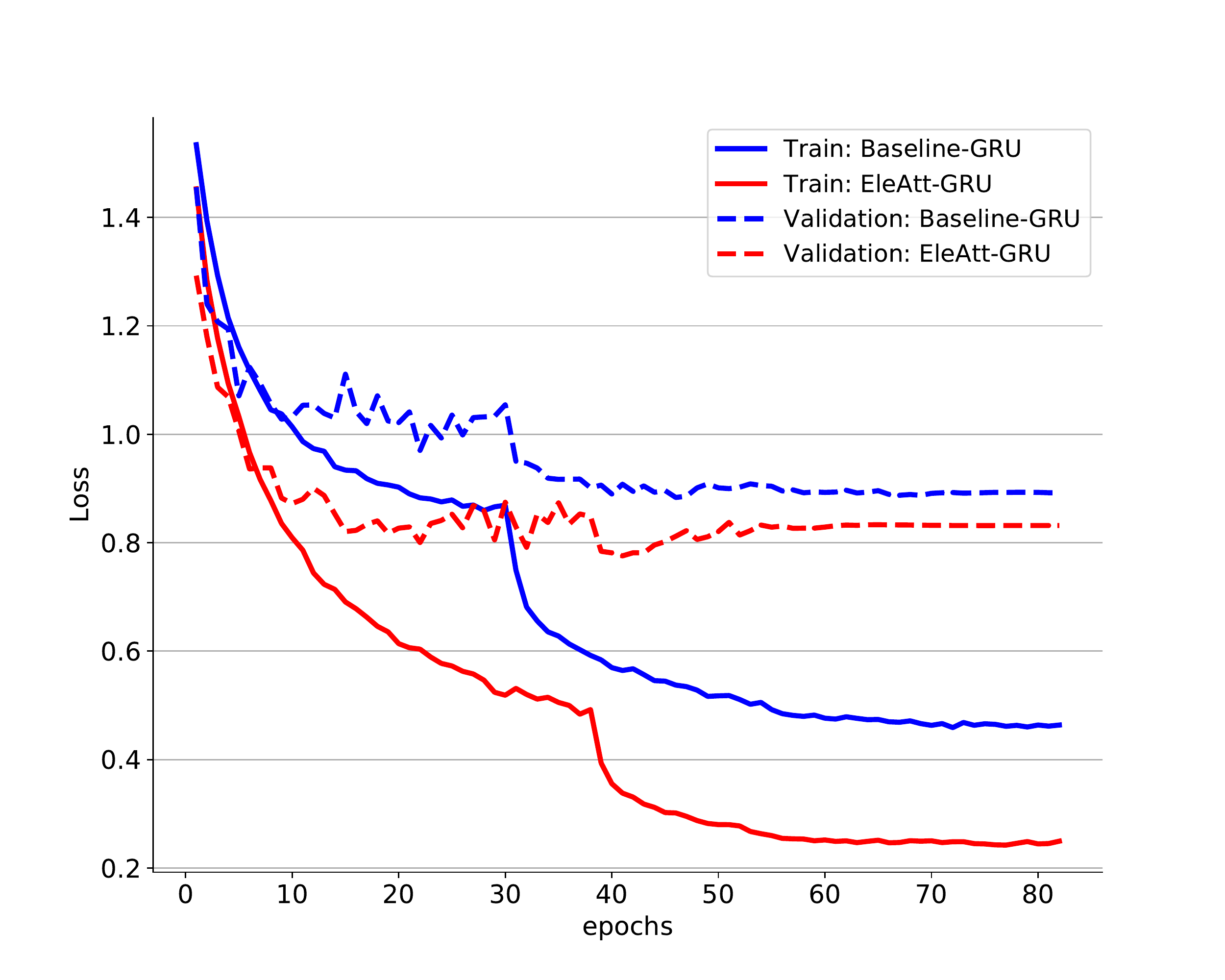}
		\caption{NTU-CV}			
		\label{subfig:CV}
	\end{subfigure}
	\caption{Loss curves during training on the NTU dataset for the proposed scheme ``EleAtt-GRU" and the baseline scheme ``Baseline-GRU".}\label{fig:curve}
\end{figure}

\subsection{Discussions}

\textbf{Convergence of Learning.} Fig. \ref{fig:curve} shows the loss curves for the training set and validation set during the training process for the proposed EleAtt-GRU and the baseline Baseline-GRU, respectively. By adding the {\EleAttGn}s, the convergence becomes faster and the final loss is much lower. EleAtt-GRU is consistently better than the baseline. The modulation of input can control the information flow of each input element adaptively and make the subsequent learning within the neurons much easier.



\noindent\textbf{Relaxing the sum-to-1 constraint on \EleAttG responses.}Unlike other works \cite{xu2015show,song2017end,liu2017global}, we do not use Softmax, which enforces the sum of attention responses to be 1, as the activation function of EleAttG. Instead, we use the Sigmoid activation function to avoid introducing  mutual influence of elements. We show the experimental comparisons between the cases with the sum-to-1 constraint ($w/ constraint$) by using Softmax, and our case without such constraint ($wo/ constraint$) by using Sigmoid in Table \ref{tab:constrain}. ``EleAttG-$n^{th}$" denotes that the $n^{th}$ GRU layer uses the GRU with EleAttG while the other layers still use the original GRU. ``Baseline" denotes the baseline scheme with three GRU layers. We can see $wo/ constraint$ always performs better than that with constraint $w/ constraint$. Adding EleAttG with constraint on the second or the third layer even decreases the accuracy by about 2.4-3.2\% in comparison with the baselines. 

\noindent\textbf{Number of parameters versus performance.} For an RNN block, the adding of an EleAttG increases the number of parameters. Taking a GRU block of $N$ neurons with the input dimension of $D$ as an example, the numbers of parameters for the original GRU block and the proposed EleAttG-GRU block are $3N(D+N+1)$, and $3N(D+N+1) +D(D+N+1)$, respectively. We calculate the computational complexity by counting the number of floating-point operations (FLOPs) including all multiplication and addition operations. At a time slot, adding attention to the layer as in \ref{equ:agru} and \ref{equ:updatedx} takes $D(D+N+1)$ multiplication operations and $D(D+N)$ addition operations. Then the complexity increases from $N(6D+6N+5)$ to $N(6D+6N+5)+D(2D+2N+1)$, which is approximately proportional to the number of parameters. 



Table \ref{tab:params} shows the effect of the number of parameters under different experimental settings on the NTU dataset. Note that ``$m$-GRU($n$)" denotes the baseline scheme which is built by $m$ GRU blocks (layers) with each layer composed of $n$ neurons. ``$m$-EleAtt-GRU(100)" denotes our scheme which includes $m$ EleAtt-GRU layers with each layer composed of 100 neurons.  We can see that the performance increases only a little when more neurons (``2-GRU(128)") or more layers (``3-GRU(100)") are used in comparison with the baseline ``2-GRU(100)". In contrast, our scheme ``2-EleAtt-GRU(100)", achieves significant gains of 3.1-4.1\%. Similar observation can be found for three-layer case. With the similar number of parameters, adding \EleAttG is much more effective than increasing the number of neurons or the number of layers.

\setlength{\tabcolsep}{3pt}
\begin{table}[t]
	\centering
	\caption{Performance comparisons about relaxing the constraint to EleAttG on the NTU dataset in terms of accuracy (\%).}
	\label{tab:constrain}
	\begin{tabular}{cccccc}
		\toprule
		Protocols           & Method              & Baseline & EleAttG-$1^{st}$ & EleAttG-$2^{nd}$ & EleAttG-$3^{rd}$ \\
		\midrule
		\multirow{2}{*}{CS} &  w/ constraint  & 75.2          & 75.0      & 72.7      & 72.0      \\
		&  wo/ constrain & 75.2          & \bf{78.7}      & \bf{77.3}      & \bf{76.4}      \\
		\midrule
		\multirow{2}{*}{CV} &  w/ constraint  & 81.5          & 83.7      & 79.1      & 78.8      \\
		&  wo/ constrain & 81.5          & \bf{84.9}      & \bf{83.5}      & \bf{82.5}     \\
		\bottomrule
	\end{tabular}
\end{table}

\setlength{\tabcolsep}{6pt}
\begin{table}[t]
	\centering
	\caption{Effect of the number of parameters  on the NTU dataset.}
	\begin{tabular}{cccc}
		\toprule
		Scheme & \# parameters & CS    & CV \\
		\midrule
		2-GRU(100) & 0.14M & 75.5  & 81.4  \\
		
		2-GRU(128) & 0.21M & 75.8 & 81.7 \\
		3-GRU(100) & 0.20M & 75.2  & 81.5  \\
		3-GRU(128) & 0.31M & 76.5  & 81.3  \\
		2-EleAtt-GRU(100) & 0.20M & 78.6  & 85.5 \\
		3-EleAtt-GRU(100) & 0.28M & 79.8  & 87.1  \\
		\bottomrule
	\end{tabular}%
	\label{tab:params}%
\end{table}%

\section{Conclusions}

In this paper, we propose to empower the neurons in recurrent neural networks to have the attentiveness capability by adding the proposed EleAttG. It can explore the varying importance of different elements of the inputs. The EleAttG is simple yet effective. Experiments show that our proposed EleAttG can be used in any RNN structures ({\it{e.g}} standard RNN, LSTM and GRU) and any layers of the multi-layer RNN networks. In addition, for both human skeleton-based and RGB-based action recognitions, EleAttG boosts performance significantly. We expect that, as a fundamental unit, the proposed EleAttG will be effective for improving many RNN-based learning tasks.

\section*{Acknowledgment}
This work is supported by  National Key Research and Development Program of China under Grant 2016YFB1001004, Natural Science Foundation of China under Grant 61773311  and Grant 61751308.
\clearpage

\bibliographystyle{splncs}
\bibliography{egbib}

\begin{thebibliography}{10}

\bibitem{lipton2015critical}
Lipton, Z.C., Berkowitz, J., Elkan, C.:
\newblock A critical review of recurrent neural networks for sequence learning.
\newblock arXiv preprint arXiv:1506.00019 (2015)

\bibitem{hochreiter1997long}
Hochreiter, S., Schmidhuber, J.:
\newblock Long short-term memory.
\newblock Neural computation \textbf{9}(8) (1997)  1735--1780

\bibitem{cho2014learning}
Cho, K., Van~Merri{\"e}nboer, B., Gulcehre, C., Bahdanau, D., Bougares, F.,
  Schwenk, H., Bengio, Y.:
\newblock Learning phrase representations using rnn encoder-decoder for
  statistical machine translation.
\newblock arXiv preprint arXiv:1406.1078 (2014)

\bibitem{du2015hierarchical}
Du, Y., Wang, W., Wang, L.:
\newblock Hierarchical recurrent neural network for skeleton based action
  recognition.
\newblock In: Computer Vision and Pattern Recognition (CVPR). (2015)
  1110--1118

\bibitem{cho14}
Cho, K., van Merri{\"{e}}nboer, B., G{\"{u}}l{\c c}ehre, {\c C}., Bahdanau, D.,
  Bougares, F., Schwenk, H., Bengio, Y.:
\newblock Learning phrase representations using rnn encoder--decoder for
  statistical machine translation.
\newblock In: EMNLP, Association for Computational Linguistics (October 2014)
  1724--1734

\bibitem{vinyals2015show}
Vinyals, O., Toshev, A., Bengio, S., Erhan, D.:
\newblock Show and tell: A neural image caption generator.
\newblock In: CVPR, IEEE (2015)  3156--3164

\bibitem{jozefowicz2015empirical}
Jozefowicz, R., Zaremba, W., Sutskever, I.:
\newblock An empirical exploration of recurrent network architectures.
\newblock In: ICML. (2015)  2342--2350

\bibitem{luong2015effective}
Luong, M.T., Pham, H., Manning, C.D.:
\newblock Effective approaches to attention-based neural machine translation.
\newblock arXiv preprint arXiv:1508.04025 (2015)

\bibitem{vaswani2017attention}
Vaswani, A., Shazeer, N., Parmar, N., Uszkoreit, J., Jones, L., Gomez, A.N.,
  Kaiser, {\L}., Polosukhin, I.:
\newblock Attention is all you need.
\newblock In: NIPS. (2017)  6000--6010

\bibitem{xu2015show}
Xu, K., Ba, J., Kiros, R., Cho, K., Courville, A., Salakhudinov, R., Zemel, R.,
  Bengio, Y.:
\newblock Show, attend and tell: Neural image caption generation with visual
  attention.
\newblock In: ICML. (2015)  2048--2057

\bibitem{li2017attentive}
Li, J., Wei, Y., Liang, X., Dong, J., Xu, T., Feng, J., Yan, S.:
\newblock Attentive contexts for object detection.
\newblock TMM \textbf{19}(5) (2017)  944--954

\bibitem{sharma2015actionattention}
Sharma, S., Kiros, R., Salakhutdinov, R.:
\newblock Action recognition using visual attention.
\newblock arXiv preprint arXiv:1511.04119 (2015)

\bibitem{wang2016hierarchical}
Wang, Y., Wang, S., Tang, J., O'Hare, N., Chang, Y., Li, B.:
\newblock Hierarchical attention network for action recognition in videos.
\newblock arXiv preprint arXiv:1607.06416 (2016)

\bibitem{LSTMblog}
Olah, C.:
\newblock Lstm.
\newblock \url{http://colah.github.io/posts/2015-08-Understanding-LSTMs/}
  (2015)

\bibitem{Shahroudy_2016_CVPR}
Shahroudy, A., Liu, J., Ng, T.T., Wang, G.:
\newblock Ntu rgb+d: A large scale dataset for 3d human activity analysis.
\newblock In: CVPR. (June 2016)

\bibitem{wang2014cross}
Wang, J., Nie, X., Xia, Y., Wu, Y., Zhu, S.C.:
\newblock Cross-view action modeling, learning and recognition.
\newblock In: CVPR. (2014)  2649--2656

\bibitem{hu2015jointly}
Hu, J.F., Zheng, W.S., Lai, J., Zhang, J.:
\newblock Jointly learning heterogeneous features for rgb-d activity
  recognition.
\newblock In: CVPR, IEEE (2015)  5344--5352

\bibitem{jhuang2013towards}
Jhuang, H., Gall, J., Zuffi, S., Schmid, C., Black, M.J.:
\newblock Towards understanding action recognition.
\newblock In: ICCV, IEEE (2013)  3192--3199

\bibitem{gers1999learning}
Gers, F.A., Schmidhuber, J., Cummins, F.:
\newblock Learning to forget: Continual prediction with lstm.
\newblock (1999)

\bibitem{gers2002learning}
Gers, F.A., Schraudolph, N.N., Schmidhuber, J.:
\newblock Learning precise timing with lstm recurrent networks.
\newblock JMLR \textbf{3}(Aug) (2002)  115--143

\bibitem{chung2014empirical}
Chung, J., Gulcehre, C., Cho, K., Bengio, Y.:
\newblock Empirical evaluation of gated recurrent neural networks on sequence
  modeling.
\newblock arXiv preprint arXiv:1412.3555 (2014)

\bibitem{veeriah2015differential}
Veeriah, V., Zhuang, N., Qi, G.J.:
\newblock Differential recurrent neural networks for action recognition.
\newblock In: ICCV, IEEE (2015)  4041--4049

\bibitem{liu2017global}
Liu, J., Wang, G., Hu, P., Duan, L.Y., Kot, A.C.:
\newblock Global context-aware attention lstm networks for 3d action
  recognition.
\newblock In: CVPR. (2017)

\bibitem{song2017end}
Song, S., Lan, C., Xing, J., Zeng, W., Liu, J.:
\newblock An end-to-end spatio-temporal attention model for human action
  recognition from skeleton data.
\newblock In: AAAI. Volume~1. (2017) ~7

\bibitem{wang2013action}
Wang, H., Schmid, C.:
\newblock Action recognition with improved trajectories.
\newblock In: ICCV. (2013)  3551--3558

\bibitem{simonyan2014two}
Simonyan, K., Zisserman, A.:
\newblock Two-stream convolutional networks for action recognition in videos.
\newblock In: NIPS. (2014)  568--576

\bibitem{donahue2015long}
Donahue, J., Anne~Hendricks, L., Guadarrama, S., Rohrbach, M., Venugopalan, S.,
  Saenko, K., Darrell, T.:
\newblock Long-term recurrent convolutional networks for visual recognition and
  description.
\newblock In: CVPR. (2015)  2625--2634

\bibitem{yue2015beyond}
Yue-Hei~Ng, J., Hausknecht, M., Vijayanarasimhan, S., Vinyals, O., Monga, R.,
  Toderici, G.:
\newblock Beyond short snippets: Deep networks for video classification.
\newblock In: CVPR. (2015)  4694--4702

\bibitem{wang2016temporal}
Wang, L., Xiong, Y., Wang, Z., Qiao, Y., Lin, D., Tang, X., Van~Gool, L.:
\newblock Temporal segment networks: Towards good practices for deep action
  recognition.
\newblock In: ECCV. (2016)  20--36

\bibitem{han2017space}
Han, F., Reily, B., Hoff, W., Zhang, H.:
\newblock Space-time representation of people based on 3d skeletal data: A
  review.
\newblock CVIU \textbf{158} (2017)  85--105

\bibitem{vemulapalli2014human}
Vemulapalli, R., Arrate, F., Chellappa, R.:
\newblock Human action recognition by representing 3d skeletons as points in a
  lie group.
\newblock In: CVPR. (2014)  588--595

\bibitem{xia2012view}
Xia, L., Chen, C.C., Aggarwal, J.:
\newblock View invariant human action recognition using histograms of 3d
  joints.
\newblock In: Computer Vision and Pattern Recognition Workshop (CVPRW), IEEE
  (2012)  20--27

\bibitem{wang2012mining}
Wang, J., Liu, Z., Wu, Y., Yuan, J.:
\newblock Mining actionlet ensemble for action recognition with depth cameras.
\newblock In: CVPR, IEEE (2012)  1290--1297

\bibitem{wang2016graph}
Wang, P., Yuan, C., Hu, W., Li, B., Zhang, Y.:
\newblock Graph based skeleton motion representation and similarity measurement
  for action recognition.
\newblock In: ECCV, Springer (2016)  370--385

\bibitem{zhu2016co}
Zhu, W., Lan, C., Xing, J., Zeng, W., Li, Y., Shen, L., Xie, X.,  et~al.:
\newblock Co-occurrence feature learning for skeleton based action recognition
  using regularized deep lstm networks.
\newblock In: AAAI. Volume~2. (2016) ~8

\bibitem{liu2016spatio}
Liu, J., Shahroudy, A., Xu, D., Wang, G.:
\newblock Spatio-temporal lstm with trust gates for 3d human action
  recognition.
\newblock In: ECCV, Springer (2016)  816--833

\bibitem{li2017adaptive}
Li, W., Wen, L., Chang, M.C., Lim, S.N., Lyu, S.:
\newblock Adaptive rnn tree for large-scale human action recognition.
\newblock In: In Computer Vision and Pattern Recognition (CVPR). (2017)
  1444--1452

\bibitem{zhang2017view}
Zhang, P., Lan, C., Xing, J., Zeng, W., Xue, J., Zheng, N.:
\newblock View adaptive recurrent neural networks for high performance human
  action recognition from skeleton data.
\newblock In: ICCV. (2017)

\bibitem{diba2017deeptemporal}
Diba, A., Sharma, V., Van~Gool, L.:
\newblock Deep temporal linear encoding networks.
\newblock In: CVPR. (2017)  2329--2338

\bibitem{chollet2015keras}
Chollet, F.:
\newblock Keras.
\newblock \url{https://github.com/fchollet/keras} (2015)

\bibitem{al2016theano}
Al-Rfou, R., Alain, G., Almahairi, A., Angermueller, C., Bahdanau, D., Ballas,
  N., Bastien, F., Bayer, J., Belikov, A., Belopolsky, A.,  et~al.:
\newblock Theano: A python framework for fast computation of mathematical
  expressions.
\newblock arXiv preprint arXiv:1605.02688 \textbf{472} (2016)  473

\bibitem{srivastava2014dropout}
Srivastava, N., Hinton, G., Krizhevsky, A., Sutskever, I., Salakhutdinov, R.:
\newblock Dropout: A simple way to prevent neural networks from overfitting.
\newblock JMLR \textbf{15}(1) (2014)  1929--1958

\bibitem{sutskever2014sequence}
Sutskever, I., Vinyals, O., Le, Q.V.:
\newblock Sequence to sequence learning with neural networks.
\newblock In: NIPS. (2014)  3104--3112

\bibitem{kingma2014adam}
Kingma, D.P., Ba, J.:
\newblock Adam: A method for stochastic optimization.
\newblock arXiv preprint arXiv:1412.6980 (2014)

\bibitem{TSNModel}
Xiong, Y.:
\newblock {TSN} model.
\newblock \url{https://github.com/yjxiong/temporal-segment-networks} (2016)

\bibitem{kuehne2011hmdb}
Kuehne, H., Jhuang, H., Garrote, E., Poggio, T., Serre, T.:
\newblock Hmdb: a large video database for human motion recognition.
\newblock In: ICCV, IEEE (2011)  2556--2563

\bibitem{he2016deep}
He, K., Zhang, X., Ren, S., Sun, J.:
\newblock Deep residual learning for image recognition.
\newblock In: CVPR. (2016)  770--778

\bibitem{ResNet50Model}
Chollet, F.:
\newblock Resnet50 model.
\newblock
  \url{https://github.com/fchollet/deep-learning-models/releases/download/v0.2/resnet50_weights_tf_dim_ordering_tf_kernels.h5}

\bibitem{evangelidis2014skeletal}
Evangelidis, G., Singh, G., Horaud, R.:
\newblock Skeletal quads: Human action recognition using joint quadruples.
\newblock In: ICPR, IEEE (2014)  4513--4518

\bibitem{ke2017new}
Ke, Q., Bennamoun, M., An, S., Sohel, F., Boussaid, F.:
\newblock A new representation of skeleton sequences for 3d action recognition.
\newblock In: CVPR, IEEE (2017)  4570--4579

\bibitem{wang2014learning}
Wang, J., Liu, Z., Wu, Y.:
\newblock Learning actionlet ensemble for 3d human action recognition.
\newblock In: Human Action Recognition with Depth Cameras.
\newblock Springer (2014)  11--40

\bibitem{hu2016real}
Hu, J.F., Zheng, W.S., Ma, L., Wang, G., Lai, J.:
\newblock Real-time rgb-d activity prediction by soft regression.
\newblock In: ECCV, Springer (2016)  280--296

\end{thebibliography}
\end{document}